\documentclass[letterpaper, 10 pt, conference]{ieeeconf}  

\IEEEoverridecommandlockouts                              

\usepackage{graphicx}
\usepackage{color}
\usepackage{soul}
\usepackage{epsfig} 
\usepackage{mathptmx} 
\usepackage{times} 
\usepackage{amsmath} 
\usepackage{amssymb}  
\usepackage{subfigure}
\usepackage{bm}
\usepackage[export]{adjustbox}
\let\labelindent\relax
\usepackage{enumitem}
\usepackage{multirow}
\usepackage{nicematrix}
\usepackage{float}
\usepackage{xspace}
\usepackage{algorithm}
\usepackage{algpseudocode}
\usepackage{xcolor,colortbl}

\newcommand{\bC}{\mathbf{C}}

\newcommand{\cH}{\mathcal{H}}

\newcommand{\cL}{\mathcal{L}}

\newcommand{\bE}{\mathbf{E}}
\newcommand{\cE}{\mathcal{E}}

\newcommand{\bI}{\mathbf{I}}

\newcommand{\bM}{\mathbf{M}}

\newcommand{\bu}{\mathbf{u}}

\newcommand{\bq}{\mathbf{q}}
\newcommand{\bR}{\mathbf{R}}

\newcommand{\bt}{\mathbf{t}}

\newcommand{\bv}{\mathbf{v}}

\newcommand{\bx}{\mathbf{x}}

\newcommand{\bK}{\mathbf{K}}

\newcommand*{\eg}{\emph{e.g.}\@\xspace}
\newcommand*{\ie}{\emph{i.e.}\@\xspace}
\newcommand*{\etal}{\emph{et al.}\@\xspace}

\definecolor{highlightgreen}{RGB}{0,255,0}
\definecolor{orbitgreen}{RGB}{89, 186, 49}
\definecolor{approachred}{RGB}{255, 31, 32}
\definecolor{bboxblue}{RGB}{0,0,255}
\definecolor{wireframegreen}{RGB}{0,200,0}
\definecolor{lineorange}{RGB}{245, 161, 66}
\definecolor{lineblue}{RGB}{66, 132, 245}

\title{\LARGE \bf
Test-Time Certifiable Self-Supervision to Bridge the Sim2Real Gap in\\Event-Based Satellite Pose Estimation}

\author{Mohsi Jawaid$^{\dagger}$, Rajat Talak$^{\ast}$, Yasir Latif$^{\dagger}$,  Luca Carlone$^{\ast}$ and Tat-Jun Chin$^{\dagger}$
\thanks{$^{\dagger}$Sentient Satellites Laboratory (SSL), Australian Institute for Machine Learning (AIML), The University of Adelaide, SA 5005, Australia.}
\thanks{$^{\ast}$Laboratory for Information and Decision Systems (LIDS), Massachusetts Institute of Technology, Cambridge, MA 02139, USA.}%
}

\begin{document}

\maketitle
\thispagestyle{empty}
\pagestyle{empty}

\begin{abstract}

Deep learning plays a critical role in vision-based satellite pose estimation. However, the scarcity of real data from the space environment means that deep models need to be trained using synthetic data, which raises the Sim2Real domain gap problem. A major cause of the Sim2Real gap are novel lighting conditions encountered during test time. Event sensors have been shown to provide some robustness against lighting variations in vision-based pose estimation. However, challenging lighting conditions due to strong directional light can still cause undesirable effects in the output of commercial off-the-shelf event sensors, such as noisy/spurious events and inhomogeneous event densities on the object. Such effects are non-trivial to simulate in software, thus leading to Sim2Real gap in the event domain. To close the Sim2Real gap in event-based satellite pose estimation, the paper proposes a test-time self-supervision scheme with a certifier module. Self-supervision is enabled by an optimisation routine that aligns a dense point cloud of the predicted satellite pose with the event data to attempt to rectify the inaccurately estimated pose. The certifier attempts to verify the corrected pose, and only certified test-time inputs are backpropagated via implicit differentiation to refine the predicted landmarks, thus improving the pose estimates and closing the Sim2Real gap. Results show that the our method outperforms established test-time adaptation schemes.

\end{abstract}

\section{Introduction}\label{sec:intro}

On-orbit servicing (OOS) includes a range of space-based activities such as refuelling, maintenance, assembly and debris removal~\cite{Cavaciuti2022in-space}. By promoting reusability, extendability and space environmental protection, OOS is crucial for sustainable space utilisation. However, the viability of OOS depends on conducting space operations safely and cost effectively, which argues for a high degree of autonomy.

Underpinning OOS are autonomous spacecraft rendezvous and docking capabilities, which require accurate and real-time estimation of the 6DoF pose of a target satellite relative to the servicer satellite in proximity range. To this end, vision-based satellite pose estimation has established its great potential~\cite{PASQUALETTOCASSINIS2019100548}, where the monocular case is a leading option due to its lower size, weight and power demands.

\begin{figure}[ht]\centering
\includegraphics[height=0.25\textwidth]{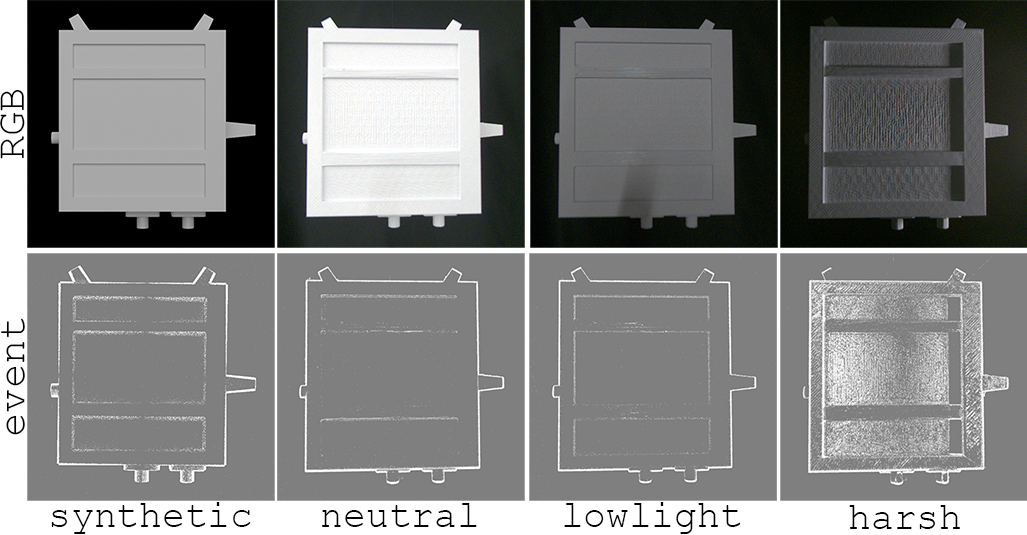}
\vspace{-1.5em}
\caption{RGB frames (top) and corresponding event frames (bottom) of a textureless 3D printed satellite-like object (see Sec.~\ref{sec:customsat} for justifications of using a textureless object) under different lighting conditions. The first column are synthetic, while the others are real. Observe that the real event frames under \texttt{neutral} and \texttt{low} are similar to the \texttt{synthetic} event frame, despite visible lighting variations in RGB. However, under \texttt{harsh} directional lighting, numerous events distributed non-uniformly on the object were generated by the event sensor. See also Sec.~\ref{sec:datasets} on sensor tuning.}
\label{fig:synthetic-vs-real}
\end{figure}

State-of-the-art vision-based satellite pose estimators employ deep learning~\cite{sharma2019spn,bospec2019,wang2022bridging,park2023spnv2}, either in an end-to-end manner or in combination with other computer vision techniques~\cite{pauly2023survey}. However, the scarcity of real data from the actual operating environment means that the deep models are often trained using synthetic data~\cite{proencca2020urso, sparkchallenge21}. This generates a simulation-to-real (Sim2Real) gap~\cite{park2021speedplus} between the training and testing domains, which raises doubts on whether the trained pose estimators can operate successfully in space.

A major cause of Sim2Real gap in vision-based satellite pose estimation are previously unseen before lighting conditions in space. It is well known that the major sources of illumination in orbit, \ie, the Sun and the Earth's albedo~\cite{beierle19variable}, can produce undesirable effects on the images of satellites in orbit, \eg, high contrast, camera over-exposure, stray light and lens flare~\cite{park2021speedplus}. Such effects are also non-trivial to simulate in software, which further contributes to the Sim2Real gap. To bridge the gap, domain adaptation methods~\cite{WANG2018135} have been applied in satellite pose estimation~\cite{sparkchallenge21,wang2022bridging,park2023spnv2}.

Another approach to mitigate the effects of novel lighting is by using optical sensors that are less susceptible to illumination variations, specifically, the event sensor~\cite{posch2010qvga}. Since event sensors asynchronously detect intensity changes, they have much higher dynamic ranges. Previous works showed that event sensors exhibited a high degree of sensing robustness across drastic lighting changes~\cite{vidal2018ultimate, rebecq2019high}. The potential benefits of event sensing to space applications have been recognised~\cite{izzo2022neuromorphic,mahlknecht2022exploring,sofiamcleod2022eccv}. In particular, Jawaid \etal~\cite{jawaid2023towards} established the viability of training a 6DoF satellite pose estimator that operates on event frames, with initial results suggesting that the Sim2Real gap for event data is smaller; see Fig.~\ref{fig:synthetic-vs-real}~(cols.~2 and 3). We will provide empirical evidence to further support the claim in Sec.~\ref{sec:results}.

\begin{figure}[t]
\subfigure[]{\includegraphics[width=0.24\textwidth,height=0.24\textwidth]{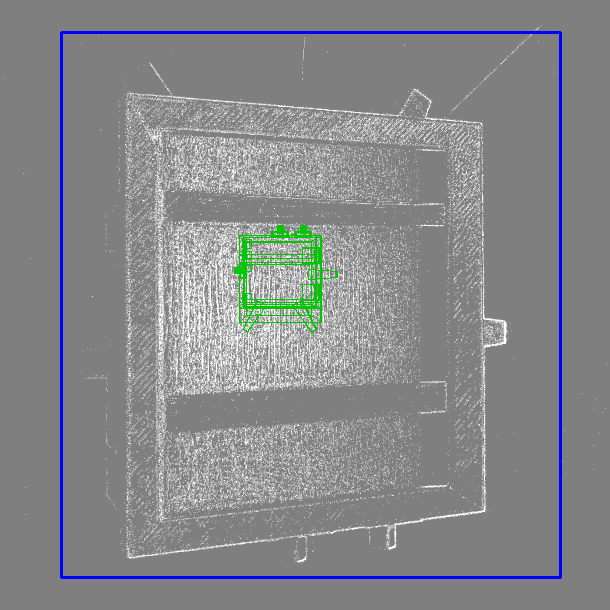}\label{fig:event-before-certifier}}
\hfill
\subfigure[]{\includegraphics[width=0.24\textwidth,height=0.24\textwidth]{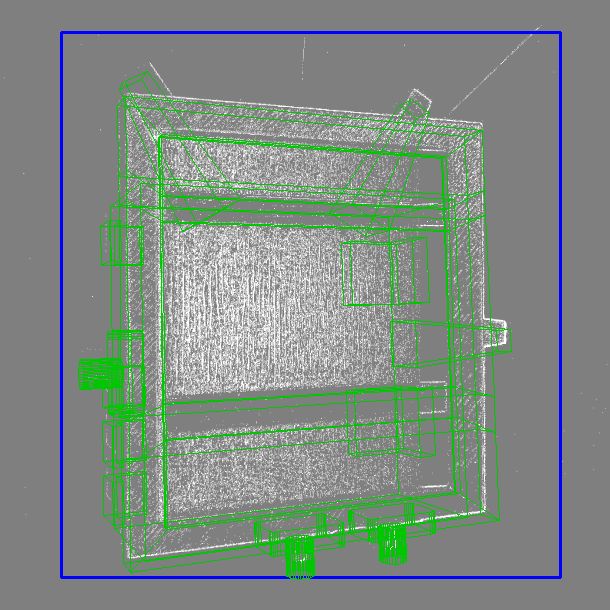}\label{fig:event-after-certifier}}
\vspace{-0.7em}
\caption{(a) Real event frame captured under harsh lighting with the object detection bounding-box (\textcolor{bboxblue}{blue}) and  projected wireframe (\textcolor{wireframegreen}{green}) of the CAD model before test-time certifiable self-supervision. (b) Same event frame as (a) but with the projected wireframe after the test-time self-supervision.}
\vspace{-0.7em}
\label{fig:before-after-adaptation}
\end{figure}

However, while event sensors offer a degree of resilience against lighting variability, harsh illumination conditions can still cause undesirable effects on commercial off-the-shelf (COTS) event sensors. In particular, non-uniform lighting due to strong directional light sources can lead to varying contrasts, which manifest as non-uniform event densities on an object; see Fig.~\ref{fig:synthetic-vs-real}~(col.~4). The pattern of real events generated are difficult to synthetically generate, which results in a domain gap for models trained on synthetic event data that did not include the effects of harsh lighting; see Fig.~\ref{fig:event-before-certifier}.

\subsection{Contributions}\label{sec:contributions2}

To deal with the Sim2Real gap in the event domain due to harsh lighting on COTS event sensors, we propose a \emph{certifiable self-supervised scheme}~\cite{talak2023certifiable,shi2023correct} to adapt the event-based pose estimator at test time. The main idea is to employ a certifier that leverages prior knowledge of the target satellite to aim to verify the correctness of estimated poses in a self-supervised manner. Only certified testing inputs are backpropagated via implicit differentiation to bridge the Sim2Real gap at test time. Departing from~\cite{talak2023certifiable,shi2023correct} who focussed on 3D point clouds, we developed a novel certifier that is optimised for the characteristics of 2D event frames (details in Sec.~\ref{sec:certifier}). See Fig.~\ref{fig:event-after-certifier} for a preview.

Compared to previous unsupervised domain adaptation methods for satellite pose estimation~\cite{wang2022bridging,park2023spnv2}, our method is source-free (whereas~\cite{wang2022bridging} requires the original labelled training data in the adaptation) and refines all learnable weights (whereas~\cite{park2023spnv2} updates only the batch norm layers in the CNN~\cite{wang2021tent}). Moreover,~\cite{wang2022bridging,park2023spnv2} need to solve auxiliary tasks (\eg, semantic segmentation) to provide additional guidance, whereas our pipeline conducts only pose estimation.

To evaluate our ideas, we developed a novel dataset for event-based satellite pose estimation. Our new dataset is an improvement over the SEENIC dataset in~\cite{jawaid2023towards}, in that real event data from emulated satellite rendezvous trajectories were captured alongside real RGB data, with both channels accompanied with common calibrated ground truth poses.

\section{Related work}

\subsection{Satellite pose estimation}

In uncooperative rendezvous and docking, the servicer spacecraft must estimate the 6DoF pose of the target spacecraft (\eg, a defunct satellite) without receiving communication or assistance from the target~\cite{fehse2014rendezvous}. Sensing modalities that have been explored for the task include radar, LiDAR and optical~\cite{goddard2010onorbit}. Monocular vision is attractive due to its lower size, weight and power demands, especially on nanosatellites which are limited in those respects~\cite{PASQUALETTOCASSINIS2019100548}.

The importance of monocular satellite pose estimation as a research topic is underlined by several datasets \cite{speed2020,park2021speedplus,proencca2020urso} and challenges \cite{spec19,spec21,sparkchallenge21}. Early adopters of deep learning for the task include Sharma \etal~\cite{sharma2018pose,sharma2019spn}. Pauly \etal~\cite{pauly2023survey} grouped existing deep learning approaches into two broad categories: hybrid modular approaches and direct end-to-end approaches. The former category of methods (\eg,~\cite{bospec2019,huo2020fast,wang2022bridging}) employ classical vision techniques (\eg, perspective-n-point solver) alongside deep learning capabilities (\eg, object detection, landmark regression), while the latter category (\eg,~\cite{sharma2018pose,sharma2019spn,park2023spnv2}) directly regresses the pose using a CNN without intermediate non-learning stages.

\subsection{Self-supervised object pose estimation}

Self-supervision has been employed to bridge the domain gap between training and testing. Wang \etal~\cite{wang2020self6d,wang2021occlusion} trained an RGB-D-based pose estimator on synthetic data and refined it with self-supervised training on real unlabelled data, with the self-supervision signal created through differentiable rendering. An iterative student-teacher scheme was proposed by Chen~\etal~\cite{chen2022sim} to bridge the Sim2Real gap. Wang \etal~\cite{wang2019normalized} extracted pose-invariant features for category-level object pose estimation of unseen object instances. Differentiable rendering of signed distance fields was utilised by Zakharov \etal~\cite{zakharov2020autolabeling}, together with normalised object coordinate spaces~\cite{wang2019normalized}, to learn 9D cuboids in a self-supervised manner. Zhang \etal~\cite{zhang2023self} proposed to jointly reconstruct the 3D shape of an object category and learn dense 2D-3D correspondences between the input image and the 3D shape, leveraging large-scale real-world object videos for training. Deng \etal~\cite{deng2020self} employed a manipulator to interact with the objects in the scene, which generates new data to fine-tune the object pose estimator. 

Departing from previous approaches that institute multiple prediction heads that solve complementary tasks, Talak \etal~\cite{talak2023certifiable,shi2023correct} advanced the certifiable self-supervision idea, where a geometric optimisation problem is solved to correct for prediction errors, while a certifier tries to verify the correction and only allows good testing samples to be backprogated via implicit differentiation to adapt the network.

As mentioned in Sec.~\ref{sec:contributions2}, domain adaptation is crucial for satellite pose estimation~\cite{wang2022bridging,park2023spnv2}. The advantages of our approach over~\cite{wang2022bridging,park2023spnv2} have been outlined in Sec.~\ref{sec:contributions2}.

\subsection{Event sensing for space applications}

While event sensing has received a significant amount of attention in robotics and AI, the potential of event sensing for space applications is only starting to be recognised~\cite{izzo2022neuromorphic}. Many of the applications relate to guidance, navigation and control (GNC) systems of spacecraft and space robots, such as star tracking and attitude estimation~\cite{chin2019star,ng2022asynchronous}, time-to-contact estimation for spacecraft landing~\cite{sikorski2021event,sofiamcleod2022eccv}, and visual odometry for planetary robots~\cite{mahlknecht2022exploring}. Azzalini~\etal~\cite{azzalini2023on} investigated the generation of synthetic events for optimal landing trajectories. To gain a better idea of the prospects of event sensing for space applications, see~\cite{izzo2022neuromorphic}.

\subsection{Event datasets for satellite pose estimation}\label{sec:datasetsurvey}

As alluded to above, Jawaid \etal~\cite{jawaid2023towards} first considered event sensing for satelite pose estimation, and a dataset was also contributed by them~\cite{elms2022seenic}. We address some shortcomings of this data, in particular we now provide equivalent RGB data as well as calibrated 6DoF object poses instead of just camera poses. Another dataset was recently proposed by Rathinam~\etal~\cite{rathinam2023spades}, however at the time of submission, their dataset has not been released publicly, hence the precise characteristics were unknown, not to mention the inability to evaluate our method on their dataset.

\section{Event-based pose estimation}

\begin{figure*}
\centering
\includegraphics[width=2\columnwidth,height=0.57\columnwidth]{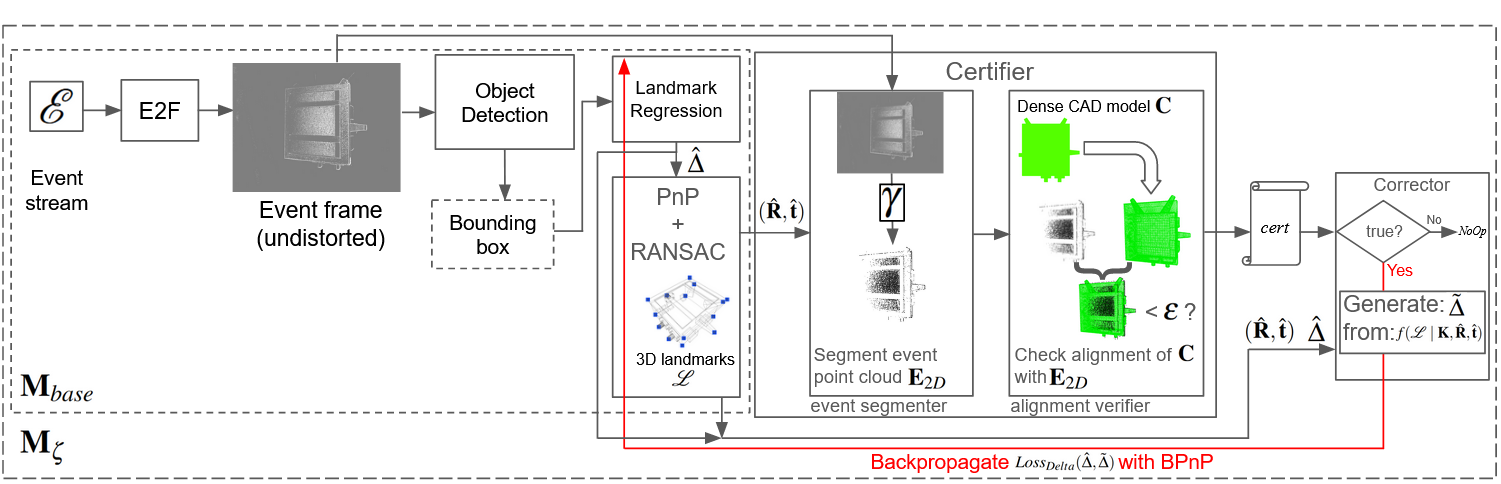}
\caption{Our test-time certifiable self-supervised for satellite pose estimation from event frames.}
\label{fig:pose_estimation_pipeline}
\vspace{-1em}
\end{figure*}

Here, we briefly describe the event-based pose estimation pipeline $\bM_{base}$~\cite{jawaid2023towards} that our work builds upon. Readers interested in more details of $\bM_{base}$ are referred to~\cite{jawaid2023towards}.

We assume that $\bM_{base}$ has been pre-trained to conduct 6DoF pose estimation based on event-frame observations of a target satellite-like object. During inference, an event stream $\cE$ is converted into event frames $\bI^{ev} = \{ I^{ev}_n \}^{N}_{n=1}$ via the E2F algorithm~\cite{jawaid2023towards} with a fixed temporal batching window $\tau$ (see Sec.~\ref{sec:impdetails} for details) where $n$ is the timestamp index for an event frame $I^{ev}_n$. Each event frame $I^{ev}_n$ is a normalised histogram of event positions where the brightest pixel indicates a pixel position with most contrast change events occurring in $\tau$. Each $I^{ev}_n$ is input to $\bM_{base}$, which conducts object detection~\cite{wu2019detectron2} that outputs a bounding box which is then passed to the landmark regression module~\cite{sun2019hrnet}. The landmark regressor produces heatmaps $\hat{\Delta} = \{ \hat{\delta}_{z} \}^{Z}_{z=1}$ of locations of $Z$ pre-selected 3D landmark positions $\cL$ on the CAD model $\bC$, which are converted into discrete 2D positions $\cL_{2D}$. A robust object pose estimation step (PnP with RANSAC) then computes the object pose $(\hat{\mathbf{R}},\hat{\bt})$ given the 2D-3D correspondences between $I^{ev}_n$ and $\bC$.

In our work, $\bM_{base}$ was trained on synthetic event frames (details in Sec.~\ref{sec:datageneration}) with only the \texttt{RandomRotation} and \texttt{RandomTranslation} augmentations~\cite{jawaid2023towards}.

\section{Test-time certifiable self-supervision}

Certifiable self-supervision for object pose estimation was first formalised in~\cite{talak2023certifiable}. Here, we describe our extension to enable test-time certifiable self-supervision to bridge the Sim2Real gap in event-based satellite pose estimation.

\subsection{Overall algorithm}\label{sec:overall_algorithm}

Our test-time certifiable self-supervision method is summarised in Alg.~\ref{alg:certifiable-self-supervision}, while the main processing pipeline $\bM_{\zeta}$ is illustrated in Fig.~\ref{fig:pose_estimation_pipeline}. Note that $\bM_{\zeta}$ subsumes $\bM_{base}$. Our method devises a certifier that leverages prior knowledge of the target satellite in the form of $\bC$ to attempt to verify the correctness of the estimated pose $(\hat{\bR},\hat{\bt})$ for a testing event frame. Certified predictions are then used by the corrector to compute the correction signal, which is backpropagated through PnP to update all learnable weights of $\bM_{base}$. We show in Sec.~\ref{sec:results} that after few epochs, the number of certified testing inputs converge to a high value, indicating that $\bM_{base}$ has been adapted to the testing (real) domain.

\begin{algorithm}[ht]\centering
\caption{Test-time certifiable self-supervision for event-based satellite pose estimation.}\label{alg:certifiable-self-supervision}
\begin{algorithmic}[1]
\Require Event-based pose estimation pipeline $\bM_{base}$, $N$ test-time event frames $\mathbf{I}^{ev}$, CAD model of satellite $\bC$, 3D landmarks $\cL$, self-training epochs $E_{self}$, self-training batch size $b_{self}$, learning rate $\alpha_{self}$, event pixel filter $\gamma$, certifier threshold $\epsilon$, threshold annealing rate $\beta$, Hausdorff distance percentile $q$.
\For{epoch $= 1,2,\dots,E_{self}$}
\For{iter $= 1,2,\dots,\lfloor N/b_{self}\rfloor$}
\State $\mathbf{I}_b \leftarrow$ Sample $b_{self}$ frames from $\mathbf{I}^{ev}$.
\For{$I~in~\bI_b$}
\State $\hat{\Delta}, \hat{\bR},\hat{\bt} \gets \bM_{base}(I)$.
\State $cert \gets \texttt{certify($I,\hat{\mathbf{R}},\hat{\bt} \mid \bC,\gamma,\epsilon,q$)}$.
\If{$cert$ is true}
    \State $\tilde{\Delta} \gets$ \texttt{corrector}$\left(\bK[\hat{\bR}~\hat{\bt}]\cL\right)$.
    \State \texttt{backprop}$\left(\bM_{base},\hat{\Delta},\tilde{\Delta}\mid \alpha_{self}\right)$.
\EndIf
\EndFor
\EndFor
\State $\epsilon \gets \beta \epsilon$.\label{step:anneal}
\EndFor
\State return $(\hat{\bR},\hat{\bt})$ predicted by $\bM_{base}$ for all frames in $\bI^{ev}$.
\end{algorithmic}
\end{algorithm}

More details of the major components in Alg.~\ref{alg:certifiable-self-supervision} will be provided in the rest of this section.

\subsection{Certification for event frames}\label{sec:certifier} 

The certifier module consists of two sub-blocks: the \emph{event segmenter} and \emph{alignment verifier}, which work together to verify whether the predicted pose $(\hat{\mathbf{R}},\hat{\bt})$ is consistent with the input event frame $I$. Details are as follows.

\subsubsection{Event segmenter}\label{sec:event-segmenter}

The input event frame $I$ is thresholded against value $\gamma$ to produce a 2D point cloud
\begin{equation}\label{eq:segmenter}
    \bE_{2D}=\{\bx \in [1,W]\times[1,H] \mid I(\bx) \ge \gamma \} := \{ \bu_p \}^{P}_{p=1},
\end{equation}
where $W$ and $H$ are respectively the width and height of image $I$. The assumption is that brighter pixels in $I$ are more likely to correspond to valid foreground of the object.

\subsubsection{Alignment verifier}\label{sec:alignment-verifier}
The CAD model $\bC$ is posed in the estimated pose $(\hat{\bR},\hat{\bt})$ via the camera projection
\begin{equation}\label{eq:cad2d_projection}
    \bC_{2D}= f(\bC \mid \bK,\hat{\mathbf{R}},\hat{\bt}) := \{ \bv_m \}^{M}_{m=1},
\end{equation}
where function $f$ executes the pinhole projection.
We then compute the Hausdorff Distance
\begin{equation}\label{eq:hausdorff_distance}
    \cH(\bE_{2D},\bC_{2D})=percentile(\mathcal{D}(\bE_{2D},\bC_{2D}), q),
\end{equation}
where $percentile(A,q)$ denotes the member value in the sorted set $A$ below which $q\%$ of members of A lie, and
\begin{align}
    \mathcal{D}(\bE_{2D},\bC_{2D}) = \left\{ \min_{m} \| \bu_p - \bv_m \|_2 \right\}^{P}_{p=1}
\end{align}
is the set of distances of the nearest neighbours of $\bE_{2D}$ in $\bC_{2D}$. Using the $q$-th largest value of $\mathcal{D}(\bE_{2D},\bC_{2D})$ enables a degree of robustness against spurious bright pixels in $I$.

\begin{figure}[ht]\centering
\includegraphics[height=0.25\textwidth]{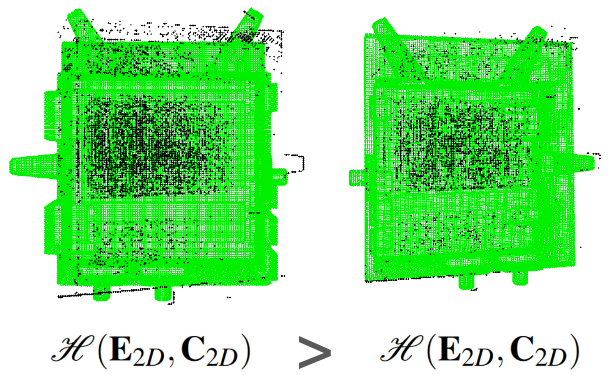}
\caption{Alignment of $\bE_{2D}$ (black) and $\bC_{2D}$ (\textcolor{wireframegreen}{green}) for a non-certified instance (left) and a certified instance (right). The hausdorff distance $\cH(\bE_{2D},\bC_{2D})$ is larger for the alignment on the left than the right.}
\label{fig:hausdorff-comparison}
\end{figure}

Finally, we say that the estimated pose $(\hat{\bR},\hat{\bt})$ for $I$ is certified (\ie, $(I,(\hat{\bR},\hat{\bt}))$ is assigned a \emph{true} certificate) if
\begin{align}
    \cH(\bE_{2D},\bC_{2D}) \le \epsilon.
\end{align}
Fig.~\ref{fig:hausdorff-comparison} depicts results corresponding to a true and false certificate. Threshold $\epsilon$ is chosen to maximise the number of valid certified instances in the test-time self-supervision batch $b_{self}$. Moreover, $\epsilon$ is progressively reduced by a rate of $\beta$ (Step~\ref{step:anneal} in Alg.~\ref{alg:certifiable-self-supervision}) to encourage convergence.

\subsection{Corrector and backpropagation}

The corrector takes the predicted pose $(\hat{\mathbf{R}},\hat{\bt})$ and projects the 3D landmarks onto $I$, \ie,
\begin{equation}\label{eq:landmarks2d_projection}
    \cL_{2D}= f(\cL \mid \bK,\hat{\mathbf{R}},\hat{\bt}) 
\end{equation}
where $\cL_{2D} := \{ \mathbf{g}_z \}^Z_{z = 1}$ are $Z$ positions in $I$ representing the projected 3D landmarks. We then create $Z$ heatmaps $\tilde{\Delta} = \{ \tilde{\delta}_z \}^{Z}_{z=1}$, where each $\tilde{\delta}_z$ is a matrix of size $W\times H$ with all $0$'s except at location $\mathbf{g}_z$ where it has the value $1$.

The differences between $\hat{\Delta}$ and $\tilde{\Delta}$ provide the correction signal to update the learnable weights in $\bM_{base}$. To this end, 
we calculate the heatmap loss $Loss_{\Delta}$ between $\tilde{\Delta}$ and $\hat{\Delta}$
\begin{equation}
    Loss_{\Delta}(\hat{\Delta},\tilde{\Delta}) = \frac{1}{Z}\sum_{z=1}^{Z}MSE(\tilde{\delta}_z, \hat{\delta}_z),
\end{equation}
which is the average mean-squared-error ($MSE$) between 
corresponding heatmaps $\tilde{\delta}_z$ and $\hat{\delta}_z$.

In order to update the landmark regressor to enable it to adapt to the testing domain, we employ BPnP~\cite{chen2020bpnp} to backpropagate $Loss_{\Delta}$ from the corrector, through the PnP solver, to all the learnable weights in the landmark regressor. The learning rate of $\alpha_{self}$ is applied in the weight updates. 

\subsection{Hyperparameter settings}\label{sec:impdetails}
\subsubsection{Offline training}
The values of the hyperparameters when offline training on synthetic data were as follows:
\begin{itemize}[leftmargin=1em]
    \item $Z = 14$ points on the surface of $\bC$ covering unique features \eg corners, end points of the antennae, prominent edges. 
    \item Training batch size $b_{offline} = 24$.
    \item Training epochs $E_{offline} = 40$.
    \item Learning-rate $\alpha_{offline} = 0.001$ in the first epoch then reduced by a factor of $0.1$ on the $25^{th}$ and $35^{th}$ epoch.
\end{itemize}
\subsubsection{Certifiable self-training}
The values of the test-time self-training hyperparameters were as follows:
\begin{itemize}[leftmargin=1em]
    \item Batch duration $\tau = 0.05$ s for the \texttt{approach} scenarios, $0.2$ s for the \texttt{orbit-slow} scenarios and $0.01$ s for the \texttt{orbit-fast} scenarios. These values were selected to optimise the signal-to-noise ratio in the event frames.
    \item Self-training batch size $b_{self} = 32$ was selected to maximise GPU usage.
    \item Self-training epochs $E_{self} = 20$.
    \item Learning-rate $\alpha_{self} = 0.001$ in the first epoch then reduced by a factor of $0.1$ on the $15^{th}$ epoch.
    \item Event pixel filter $\gamma = 0.9$.
    \item Epsilon $\epsilon = 100$ in the first epoch of the self-training and was reduced by $\beta = 0.975$ at each subsequent epoch.
    \item Hausdorff distance percentile $q = 0.9997$.
\end{itemize}

\section{Experimental setup}

\subsection{3D model of target satellite}\label{sec:customsat}

We designed a 3D model of a satellite-like object (see Fig.~\ref{fig:synthetic-vs-real}) that is geometrically similar to satellites employed in previous satellite pose estimation works~\cite{park2021speedplus,speed2020}.

In order to focus on the effects of lighting variations on the Sim2Real gap, the following design choices were made:
\begin{itemize}[leftmargin=1em]
    \item The CAD model was textureless to preclude as much as possible the contributions of discrepancies in the digital and physical textures to the Sim2Real gap.
    \item The CAD model was specified to be of a homogeneous material which was hand tuned in Blender~\cite{blender} to closely match the reflectance of 3D printed PLA surfaces.
\end{itemize}

\subsection{Synthetic RGB and event datasets for training}\label{sec:datageneration}

The model in Sec.~\ref{sec:customsat} was rendered using Blender. The model was moved smoothly towards the virtual camera while rotating the camera about the y and z axes simultaneously. Then, the camera was kept static and the model was rotated individually about the x, y and z axes and simultaneously about all pairs of the x, y and z axes. This yielded $2,520$ sequential rendered RGB frames with ground-truth object pose for each frame. Successive poses differ by $1^\circ$ on at most 2 axes. A lighting condition was created with 2 identical lights of $2$ m radius positioned at distances of $2$ m and $-2$ m in the world z axis, and $1$ m in the world y axis. This  created even diffused lighting around the object which casted minimal shadows; see Fig.~\ref{fig:synthetic-vs-real}~(top left). To generate the synthetic event frames, we employed V2E~\cite{v2e2012katz} with the sequential RGB frames as input, and set a matching framerate and exposure duration (\eg, 10 fps and 0.1 s respectively). This ensured that we obtain a one-to-one pairing of the synthetic RGB frames and event frames; see Fig.~\ref{fig:synthetic-vs-real}~(col 1).

\subsection{Real RGB and event datasets for testing}\label{sec:datasets}

\subsubsection{Model fabrication}

We 3D printed the model in Sec.~\ref{sec:customsat} to the physical dims.~of $26$ cm $\times$ $25$ cm $\times$ 10 cm.

\subsubsection{Sensors}

We employed a Prophesee EVK1 event camera with a $1280\times720$ pixel event sensor and a Basler Ace~2 conventional camera with a $1920\times1200$ pixel Sony IMX392 optical sensor. The same Arducam 4--12 mm Varifocal C-Mount lens was used for both cameras, with the focal length tuned separately to ensure similar pixel coverage of the target object given the object physical distance in our laboratory.

\subsubsection{Lighting conditions}

The printed 3D model was imaged under 3 lighting conditions:
\begin{itemize}[leftmargin=1em]
    \item \texttt{neutral}: illuminated with evenly-lit ambient lighting.
    \item \texttt{low}: illuminated with 1 weak light source.
    \item \texttt{harsh}: illuminated with 1 ring light close to the object. 
\end{itemize}
See Fig.~\ref{fig:synthetic-vs-real} (cols 2--4). \texttt{low} provided a dimly lit environment. \texttt{harsh} was challenging due to strong directional light, high sensor noise, occlusions due to shadows, and lens flares when the light source was within the field-of-view of the camera. 

\subsubsection{Trajectories}

Following~\cite{jawaid2023towards}, two trajectories named \texttt{approach} and \texttt{orbit} were executed using a Universal Robots UR5 robot manipulator. \texttt{approach} simulated a docking manoeuvre where the camera linearly approached the satellite, which allowed to evaluate translation errors. The \texttt{orbit} trajectory simulated the chaser flying by the target, which allowed to evaluate rotation errors. See~\cite[Fig.~5]{jawaid2023towards} for an illustration of the trajectories. Both of the trajectories were executed in \texttt{fast} and \texttt{slow} speeds, where \texttt{fast} was $5\times$ higher speed and acceleration than \texttt{slow}. The different speeds allowed to investigate the effects of motion blur.

\begin{table}[H]
\centering
\begin{NiceTabular}{llccc}[hvlines,rules/color=[gray]{0.3}]
Trajectory & Scene & No.events & No.poses & Duration (s) \\
\Block{6-1}{approach} 
& harsh-fast & 5842584 & 41 & 1.6 \\
& harsh-slow & 6463394 & 90 & 3.56 \\
& lowlight-fast & 1399075 & 41 & 1.6 \\
& lowlight-slow & 1578355 & 90 & 3.56 \\
& neutral-fast & 1475790 & 41 & 1.6 \\
& neutral-slow & 1814205 & 91 & 3.6 \\
\Block{6-1}{orbit} 
& harsh-fast & 13465922 & 68 & 2.68 \\
& harsh-slow & 15056827 & 165 & 6.56 \\
& lowlight-fast & 2126339 & 68 & 2.68 \\
& lowlight-slow & 2119757 & 165 & 6.56 \\
& neutral-fast & 2171772 & 68 & 2.68 \\
& neutral-slow & 2568447 & 165 & 6.56 \\
\end{NiceTabular}
\caption{Summary of the real event data.}
\vspace{-2em}
\label{tab:event_data_summary}
\end{table}

\begin{table}[H]
\centering
\begin{NiceTabular}{llccc}[hvlines,rules/color=[gray]{0.3}]
\Block{2-1}{Trajectory/\\RGB-setting} & \Block{2-1}{Scene} & \Block{2-1}{No.\\frames} & \Block{2-1}{No.\\poses} & \Block{2-1}{Duration (s)} \\
& & & & \\
\Block{6-1}{approach/\\exposure\\priority} 
& harsh-fast & 34 & 34 & 1.651 \\
& harsh-slow & 72 & 72 & 3.552 \\
& lowlight-fast & 33 & 33 & 1.601 \\
& lowlight-slow & 73 & 73 & 3.601 \\
& neutral-fast & 33 & 33 & 1.601 \\
& neutral-slow & 73 & 73 & 3.601 \\
\Block{6-1}{orbit/\\exposure\\priority} 
& harsh-fast & 55 & 55 & 2.752 \\
& harsh-slow & 138 & 138 & 6.854 \\
& lowlight-fast & 57 & 57 & 2.802 \\
& lowlight-slow & 138 & 138 & 6.855 \\
& neutral-fast & 57 & 57 & 2.802 \\
& neutral-slow & 139 & 139 & 6.904 \\
\Block{6-1}{approach/\\motion\\priority} 
& harsh-fast & 97 & 97 & 1.603 \\
& harsh-slow & 217 & 217 & 3.607 \\
& lowlight-fast & 96 & 96 & 1.653 \\
& lowlight-slow & 218 & 218 & 3.623 \\
& neutral-fast & 97 & 97 & 1.603 \\
& neutral-slow & 217 & 217 & 3.606 \\
\Block{6-1}{orbit/\\motion\\priority} 
& harsh-fast & 169 & 169 & 2.805 \\
& harsh-slow & 414 & 414 & 6.900 \\
& lowlight-fast & 169 & 169 & 2.805 \\
& lowlight-slow & 414 & 414 & 6.899 \\
& neutral-fast & 170 & 170 & 2.821 \\
& neutral-slow & 408 & 408 & 6.882 \\
\end{NiceTabular}
\caption{Summary of the real RGB data.}
\vspace{-5mm}
\label{tab:rgb_data_summary}
\end{table}

\subsubsection{Data capture}

With the event camera we captured an event stream $\cE$ for each motion and light setting, where an event is a tuple with image coordinates, polarity and timestamp indicating intensity change. Note that tuning the event sensor bias settings did not resolve the undesirable outputs under the \texttt{harsh} condition, while tuning for the \texttt{low} condition would produce many spurious events making the data unusable. Thus, we utilised only the default event sensor settings. Adaptive tuning of event sensors is an open problem~\cite{gracca2023shining} and is beyond the scope of this paper.

\begin{figure}[H]\centering
\includegraphics[width=0.48\textwidth]{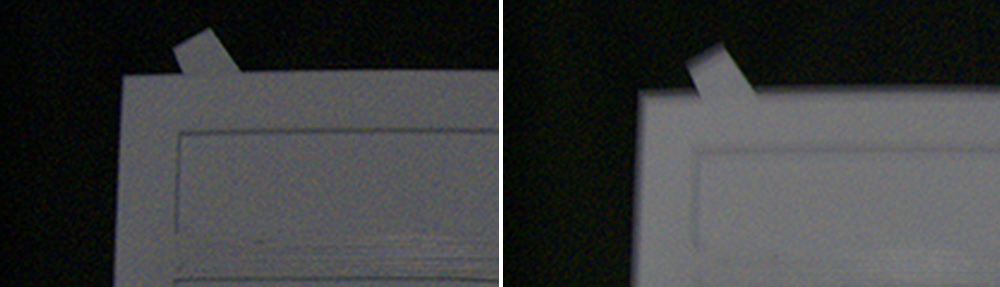}
\caption{Zoomed in view to demonstrate the difference between RGB frames captured with \texttt{motionpriority} and \texttt{exposurepriority} settings.}
\vspace{-2mm}
\label{fig:motionpriority-vs-exposurepriority}
\end{figure}

With the conventional camera we captured $J$ RGB frames $\{ I^{rgb}_j \}^{J}_{j=1}$ where $j$ is the timestamp index for a captured RGB frame $I^{rgb}_j$. To examine the behaviour of the optical sensor under lighting variations, two exposure settings were used: \texttt{motionpriority} which had a low shutter speed resulting in $60$ fps, and \texttt{exposurepriority} which optimised the exposure to the \texttt{low} lighting condition resulting in $20$ fps. Fig.~\ref{fig:motionpriority-vs-exposurepriority} shows images under the two settings.

The robot arm was polled at 60 Hz to obtain the robot base to gripper pose $(R^{b2g}_k,\bt^{b2g}_k)$, where $k$ is the robot timestamp.

\subsubsection{Intrinsic calibration}

The camera intrinsics and lens radial distortion parameters were estimated with the aid of a checkerboard pattern displayed on a screen (using respectively the Metavision SDK~\cite{MetavisionSDKDocs} for the event camera and OpenCV~\cite{opencv_library} for the frame-based camera).

\vspace{-1em}

\begin{figure}[H]
\subfigure[]{\includegraphics[width=0.22\textwidth,height=0.22\textwidth]{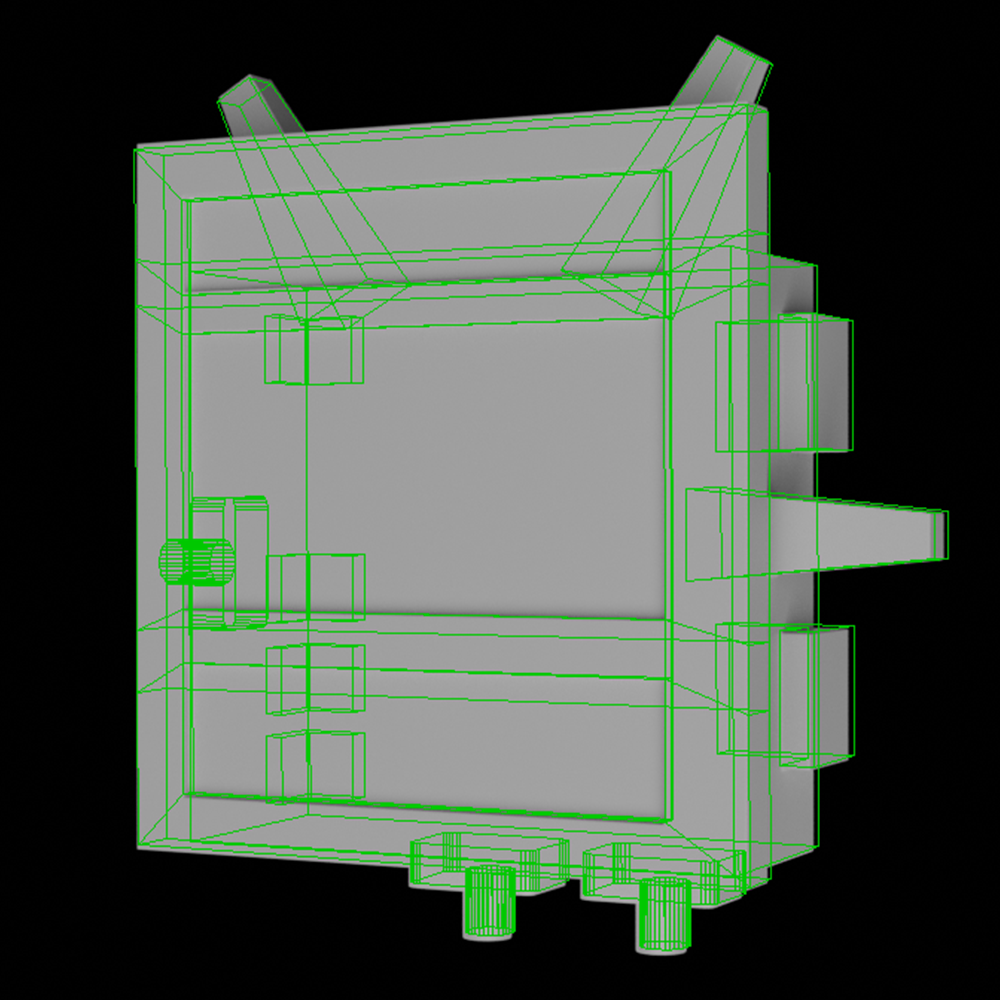}
\label{fig:synthetic_gt_rgb}}
\hfill
\subfigure[]{\includegraphics[width=0.22\textwidth,height=0.22\textwidth]{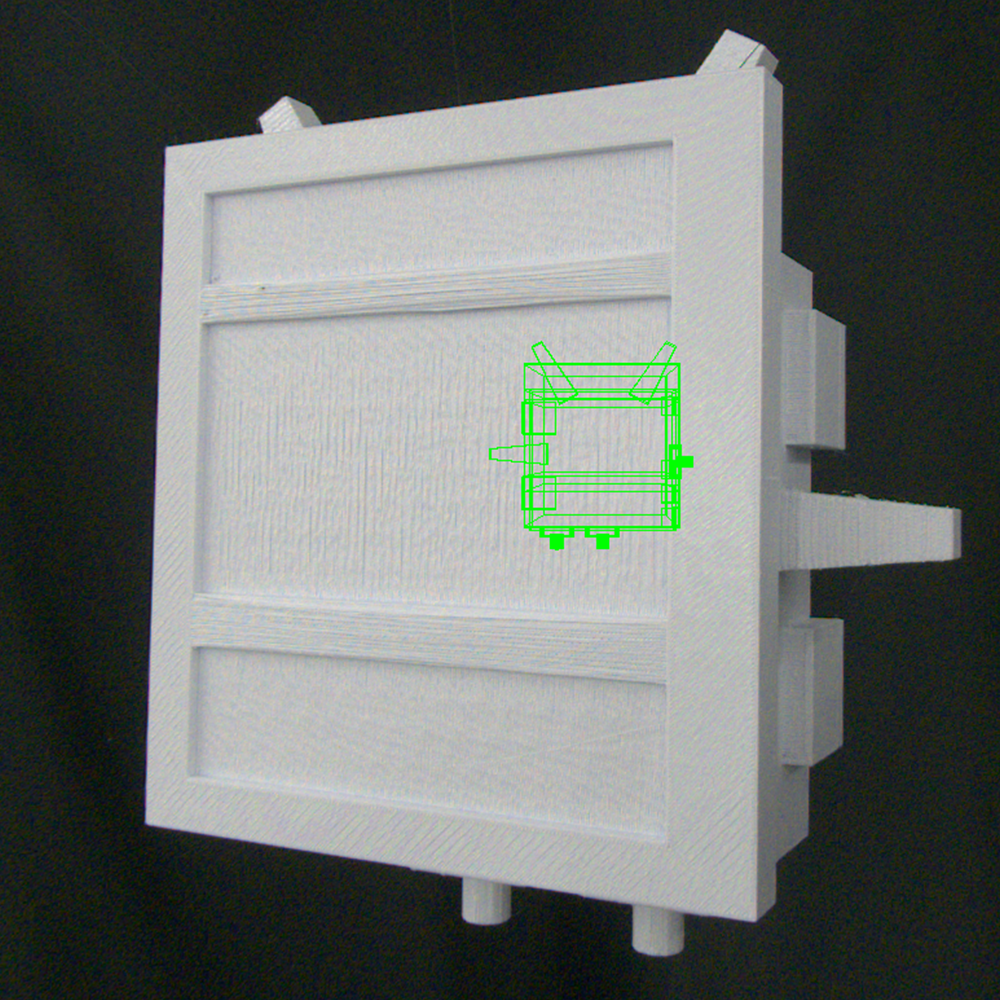}
\label{fig:real_pred_rgb}}

\vspace{-0.7em}

\subfigure[]{\includegraphics[width=0.22\textwidth,height=0.22\textwidth]{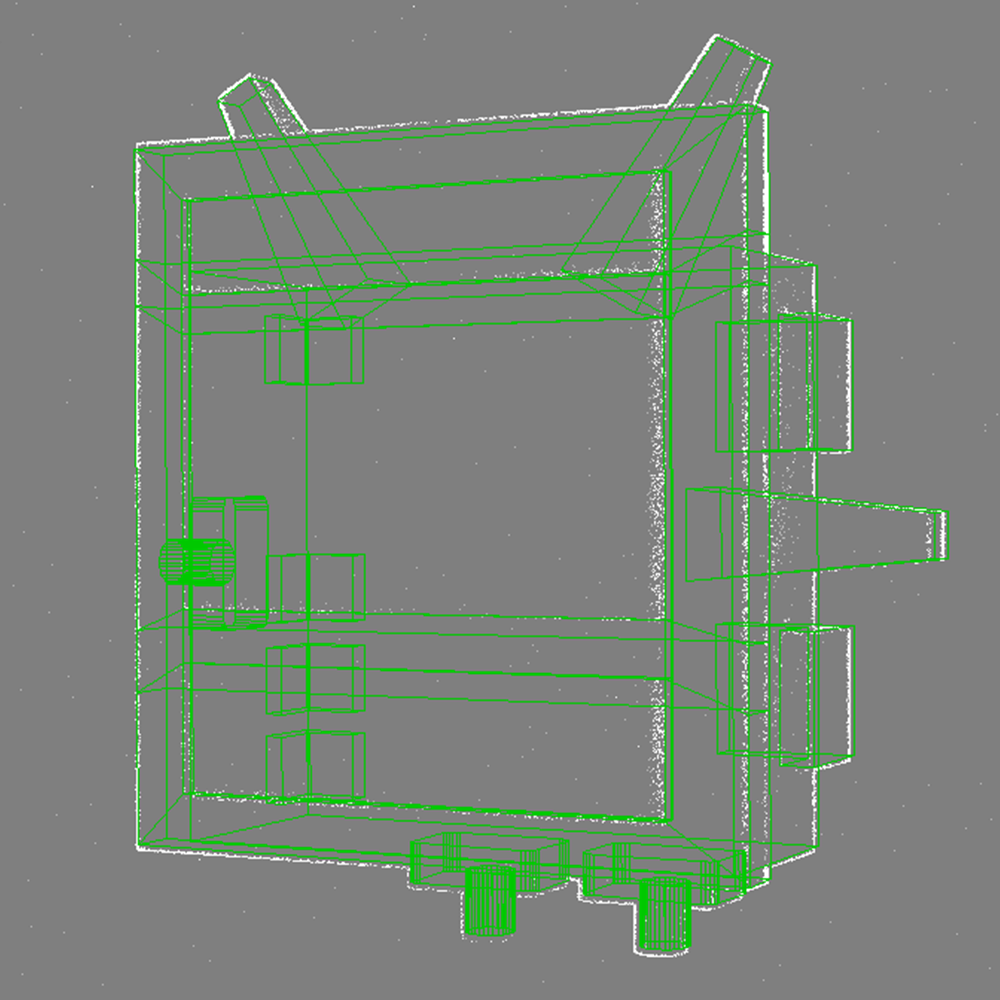}\label{fig:synthetic_gt_event}}
\hfill
\subfigure[]{\includegraphics[width=0.22\textwidth,height=0.22\textwidth]{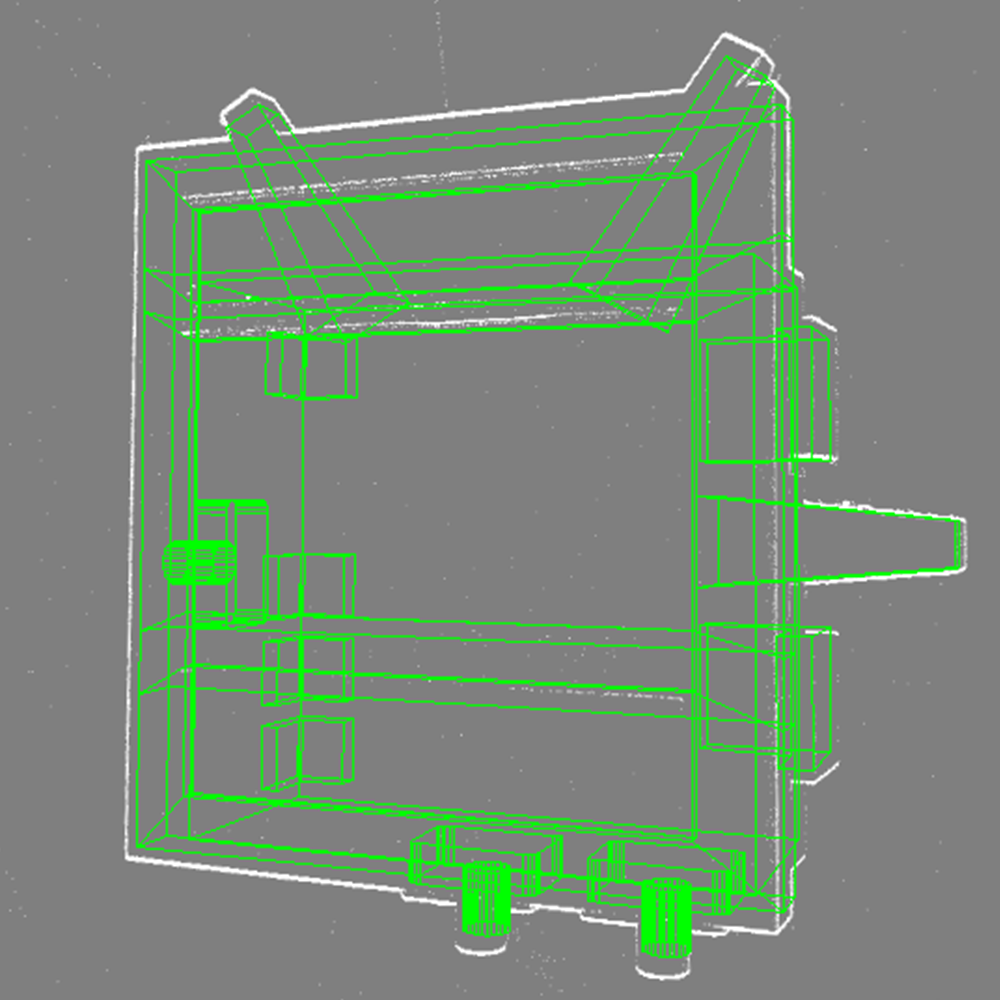}
\label{fig:real_pred_event}}

\vspace{-1em}

\caption{(a) Rendered frame of a satellite under normal lighting, with projected wireframe model. (b) Real image of a 3D model of the satellite under neutral lighting; to the human eye (a) and (b) look similar. (c) Event frame corresponding to (a), generated using V2E~\cite{hu2021v2e}. (d) Real event frame corresponding to (b). We demonstrate that the domain gap between (c) and (d) is lower than that between (a) and (b).}
\label{fig:rgb_vs_event_domaingap}
\vspace{-1em}
\end{figure}

\subsubsection{Ground-truth from calibration}
We obtain the gripper to camera constant rigid transform $(\mathbf{R}^{g2c},\bt^{g2c})$ and the base to checkerboard constant transform $(\mathbf{R}^{b2w},\bt^{b2w})$ by passing in base to gripper poses $(\mathbf{R}^{b2g}_k,\bt^{b2g}_k)$ polled from the robot arm  to OpenCV's \texttt{calibrateRobotWorldHandEye} with the estimated transforms from the checkerboard to the camera $(\mathbf{R}^{w2c}_k,\bt^{w2c}_k)$ from \texttt{solvePnP}. A satellite to camera transformation $(\mathbf{R}^{s2c}_k,\bt^{s2c}_k)$ is computed by manually annotating the landmarks in one frame under neutral lighting. Finally we obtain the satellite to camera poses $\{(\mathbf{R}_k,\bt_k)\}_{k=1}^K$ as
\begin{equation*}\label{eq:satellite2cam}
    T(\mathbf{R}_k,\bt_k)=T(\mathbf{R}^{b2s}_k,\bt^{b2s}_k)^{-1}T(\mathbf{R}^{b2g}_k,\bt^{b2g}_k)T(\mathbf{R}^{g2c},\bt^{g2c}),
\end{equation*}
where $T(\bR,\bt)$ indicates the $4 \times 4$ matrix form of the rigid transformation $(\bR,\bt)$, and
\begin{equation*}\label{eq:base2satellite}T(\mathbf{R}^{b2s}_k,\bt^{b2s}_k)=T(\mathbf{R}^{b2g}_k,\bt^{b2g}_k)T(\mathbf{R}^{g2c},\bt^{g2c})T(\mathbf{R}^{s2c}_k,\bt^{s2c}_k)^{-1}.
\end{equation*}

\subsubsection{Dataset statistics}

Tabs.~\ref{tab:event_data_summary} and~\ref{tab:rgb_data_summary} summarise the real event and RGB datasets produced\footnote{Our datasets will be released at: https://github.com/mohsij/space-opera.}.

\section{Results}\label{sec:results}

We evaluated our approach on the datasets described in Sec.~\ref{sec:customsat}. As alluded to in Sec.~\ref{sec:datasetsurvey}, SEENIC~\cite{elms2022seenic} was unsuitable since it did not have simultaneous event and RGB data, while SPADES~\cite{rathinam2023spades} was not publicly available yet.

\subsection{Details of evaluation metrics}\label{sec:evaluationmetrics}

Following Sec.~\ref{sec:datasets} (Data capture), we converted each real event stream $\cE$ to event frames $\{I^{ev}_n\}_{n=1}^N$. Each ground truth object pose $(\mathbf{R}_k,\bt_k)$ was associated with an event frame $I^{ev}_n$ nearest in timestamp to obtain a sequence $\bI^{ev}_K=\{I^{ev}_k\}_{k=1}^K \subset \bI^{ev}$ synchronised with the ground truth poses. In a similar way, a sequence of RGB frames $\bI^{rgb}_K=\{I^{rgb}_k\}_{k=1}^K \subset \bI^{rgb}$ with ground truth was obtained. Given poses $\{(\hat{\mathbf{R}}_k,\hat{\bt}_k)\}_{k=1}^K$ estimated from $\bI^{ev}_K$ or $\bI^{rgb}_K$, following~\cite{speed2020, park2021speedplus} we compare them to the ground truth poses using 
\begin{align*}
\phi_k = \frac{\left\| \bt_k - \hat{\bt}_k \right\|_2}{\left\|\bt_k\right\|_2},~~~~\psi_k =  2\arccos \left( | <\bq_k, \hat{\bq}_k> | \right),
\end{align*}
which resp.~measure translation error and rotation error, and $\bq_k$ and $\hat{\bq}_{k}$ are the quaternion form of $\mathbf{R}_k$ and $\hat{\mathbf{R}}_k$. The overall error for an event stream $\cE$ or RGB sequence $\bI^{rgb}$ are
\begin{align*}
\Phi = mean(\{\phi_k\}_{k=1}^K),~~\Psi = mean(\{\psi_k\}_{k=1}^K).
\end{align*}

\subsection{Is the domain gap smaller in event data?}\label{sec:results_rgb_vs_events}
In order to test the lighting Sim2Real domain gap in event data, we compared it against conventional RGB data. We trained the pose estimation pipeline $\bM_{base}$ separately on \texttt{synthetic} RGB and event data described in Sec.~\ref{sec:datageneration} and then run inference of $\bM_{base}$ trained on \texttt{synthetic} RGB on real RGB data and $\bM_{base}$ trained on \texttt{synthetic} event on real event data. $\Phi$ and $\Psi$ are computed for each real event stream $\cE$ and real RGB sequence $\bI^{rgb}$ which are used as metrics to quantify the domain gap between the \texttt{synthetic} training data and real testing data. Table~\ref{tab:results_rgb_vs_events} shows the values of $\Phi$ and $\Psi$ for each permutation of trajectory, lighting and speed setting. From this table we can clearly see that the $\bM_{base}$ trained on \texttt{synthetic} event data performs well when tested on real event data whereas, the $\bM_{base}$ trained on \texttt{synthetic} RGB data has a notably wider domain gap given the much larger error values with the exception of $3$ scenes where the mean value is skewed by few outlying frames. From this we can conclude that indeed the Sim2Real gap is smaller in event data as compared to RGB data. A qualitative result of the same experiment is shown in Fig.~\ref{fig:rgb_vs_event_domaingap}.

While the Sim2Real gap is smaller in the event domain, there was still appreciable error particularly on sequences with \texttt{harsh} lighting. The next experiment will investigate the effectiveness of the proposed certifiable self-supervision method in bridging the remaining Sim2Real gap.
\begin{table}[H]
\centering
\begin{NiceTabular}{llcccc}[hvlines,rules/color=[gray]{0.3}]
\Block{2-1}{Trajectory/\\RGB-setting} & \Block{2-1}{Scene} & \Block{1-2}{$\Phi$} & & \Block{1-2}{$\Psi$ (rad)} &\\
& & RGB & Event & RGB & Event \\
\Block{6-1}{approach/\\exposure\\priority}
& harsh-fast & 6.544 & \cellcolor{green}1.580 & 2.900 & \cellcolor{green}0.182 \\
& harsh-slow & 2.545 & \cellcolor{green}0.040 & 2.977 & \cellcolor{green}0.054 \\
& low-fast & 28.446 & \cellcolor{green}0.035 & 2.846 & \cellcolor{green}0.062 \\
& low-slow & 62.276 & \cellcolor{green}0.035 & 2.862 & \cellcolor{green}0.061 \\
& neutral-fast & 1.703 & \cellcolor{green}0.041 & 3.084 & \cellcolor{green}0.060 \\
& neutral-slow & 1.267 & \cellcolor{green}0.036 & 3.084 & \cellcolor{green}0.053 \\
\Block{6-1}{orbit/\\exposure\\priority}
& harsh-fast & \cellcolor{green}1.635 & 10.074 & 2.509 & \cellcolor{green}1.004 \\
& harsh-slow & \cellcolor{green}1.600 & 42.112 & 2.827 & \cellcolor{green}0.907 \\
& low-fast & 125.378 & \cellcolor{green}39.917 & 2.661 & \cellcolor{green}0.727 \\
& low-slow & 29.354 & \cellcolor{green}8.909 & 2.713 & \cellcolor{green}0.814 \\
& neutral-fast & 1.405 & \cellcolor{green}0.120 & 2.843 & \cellcolor{green}0.253 \\
& neutral-slow & \cellcolor{green}1.715 & 1.936 & 2.870 & \cellcolor{green}0.200 \\
\Block{6-1}{approach/\\motion\\priority}
& harsh-fast & 41.749 & \cellcolor{green}1.580 & 2.758 & \cellcolor{green}0.182 \\
& harsh-slow & 58.674 & \cellcolor{green}0.040 & 2.727 & \cellcolor{green}0.054 \\
& low-fast & 18.292 & \cellcolor{green}0.035 & 2.934 & \cellcolor{green}0.062 \\
& low-slow & 10.344 & \cellcolor{green}0.035 & 3.056 & \cellcolor{green}0.061 \\
& neutral-fast & 1.764 & \cellcolor{green}0.041 & 3.060 & \cellcolor{green}0.060 \\
& neutral-slow & 56.705 & \cellcolor{green}0.036 & 3.078 & \cellcolor{green}0.053 \\
\Block{6-1}{orbit/\\motion\\priority}
& harsh-fast & 537.502 & \cellcolor{green}10.074 & 2.627 & \cellcolor{green}1.004 \\
& harsh-slow & 471.970 & \cellcolor{green}42.112 & 2.611 & \cellcolor{green}0.907 \\
& low-fast & 62.245 & \cellcolor{green}39.917 & 2.570 & \cellcolor{green}0.727 \\
& low-slow & 37.387 & \cellcolor{green}8.909 & 2.612 & \cellcolor{green}0.814 \\
& neutral-fast & 2.116 & \cellcolor{green}0.120 & 2.839 & \cellcolor{green}0.253 \\
& neutral-slow & 43.310 & \cellcolor{green}1.936 & 2.868 & \cellcolor{green}0.200 \\
\end{NiceTabular}
\caption{Comparison of the errors in the RGB and event domains.}
\vspace{-1.5em}
\label{tab:results_rgb_vs_events}
\end{table}

\vspace{-1em}

\subsection{Is test-time certifiable self-supervision effective?}\label{sec:certifier_ablation_results}
\begin{table}[H]
\vspace{-1.5em}
\centering
\begin{NiceTabular}{llcccc}[hvlines,rules/color=[gray]{0.3}]
\Block{2-1}{} & \Block{2-1}{} & \Block{2-2}{$\Phi$} & & \Block{2-2}{$\Psi$ (rad)} &\\
& & & & & \\
Trajectory & Scene & $\bM_{base}$ & $\bM_{\zeta}$ & $\bM_{base}$ & $\bM_{\zeta}$ \\
\Block{6-1}{approach}
& harsh-fast & \cellcolor{green}1.580 & 21.529 & \cellcolor{green}0.059 & 0.098 \\
& harsh-slow & 0.040 & \cellcolor{green}0.021 & \cellcolor{green}0.054 & 0.085 \\
& low-fast & \cellcolor{green}0.035 & 0.056 & 0.062 & \cellcolor{green}0.058 \\
& low-slow & 0.035 & \cellcolor{green}0.024 & \cellcolor{green}0.061 & 0.106 \\
& neutral-fast & \cellcolor{green}0.041 & 0.049 & 0.060 & \cellcolor{green}0.053 \\
& neutral-slow & 0.036 & \cellcolor{green}0.012 & \cellcolor{green}0.053 & 0.110 \\
\Block{6-1}{orbit}
& harsh-fast & 10.074 & \cellcolor{green}0.021 & 0.210 & \cellcolor{green}0.055 \\
& harsh-slow & \cellcolor{green}42.112 & 79.796 & 0.264 & \cellcolor{green}0.143 \\
& low-fast & 39.917 & \cellcolor{green}0.030 & 0.085 & \cellcolor{green}0.045 \\
& low-slow & 8.909 & \cellcolor{green}0.028 & 0.152 & \cellcolor{green}0.061 \\
& neutral-fast & 0.120 & \cellcolor{green}0.019 & 0.069 & \cellcolor{green}0.039 \\
& neutral-slow & 1.936 & \cellcolor{green}0.031 & \cellcolor{green}0.081 & 0.224 \\
\end{NiceTabular}
\caption{Ablation study for certifiable self-training.}
\vspace{-2em}
\label{tab:certifier_ablation_results}
\end{table}

\begin{figure}[H]\centering
\includegraphics[height=0.25\textwidth]{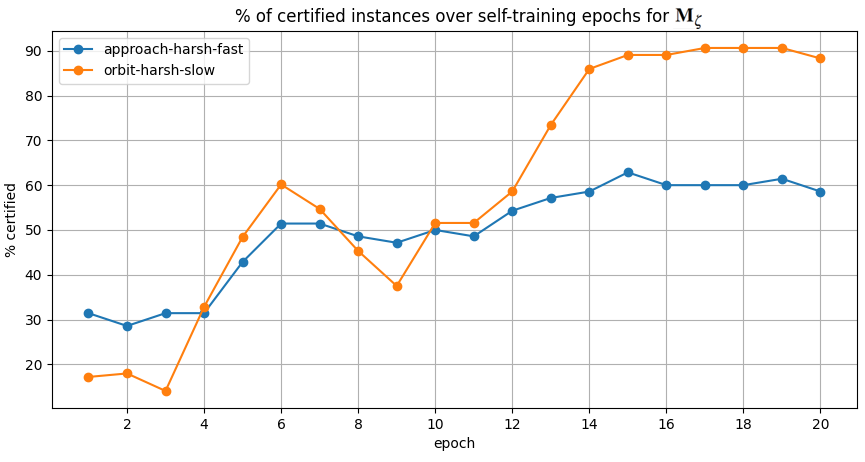}
\vspace{-1.5em}
\caption{Plot for two representative scenes showing percentage of instances where the certification is true across the self-supervision epochs.}
\label{fig:percent-certified}
\vspace{-1em}
\end{figure}

We evaluate the difference between pose estimation pipelines $\bM_{base}$ and $\bM_{\zeta}$ to determine whether the test-time certifiable self-supervision is an effective method to close the remaining domain gap in event data. We train $\bM_{base}$ and $\bM_{\zeta}$ on \texttt{synthetic} event data and then conduct inference on both pipelines to obtain the overall translation and rotation errors $\Phi$ and $\Psi$ respectively. These errors shown in Table~\ref{tab:certifier_ablation_results} indicate that the certifiable self-supervision routine definitely improves performance to estimate the translations and although the rotation scores $\Psi$ seem to suggest that $\bM_{base}$ has equivalent performance in determining the orientation of the satellite, we remind readers as mentioned in Sec.~\ref{sec:datasets} (Trajectories) that the \texttt{orbit} scenes are intended to effectively evaluate the pipeline's ability to estimate the rotation as consecutive poses in the \texttt{orbit} scenes display significant rotation changes. This trajectory provides a clearer idea of how well the pipeline performs in estimating rotations. From these results we can see that our test-time certifiable self-supervision for event data is effective. Fig.~\ref{fig:percent-certified} is also provided to demonstrate that over the self-supervision routine, the number of certified instances (for initial $\epsilon=100$) increases as the landmark regressor weights are updated. This indicates that the Sim2Real gap has been alleviated.

\subsection{Does the certifiable pipeline surpass other TTA methods?}\label{sec:results_spnv2}
\begin{table}[H]
\vspace{-1em}
\centering
\begin{NiceTabular}{llcccc}[hvlines,rules/color=[gray]{0.3}]
\Block{2-1}{} & \Block{2-1}{} & \Block{2-2}{$\Phi$} & & \Block{2-2}{$\Psi$ (rad)} &\\
& & & & & \\
Trajectory & Scene & SPNv2 & $\bM_{\zeta}$ & SPNv2 & $\bM_{\zeta}$ \\
\Block{6-1}{approach}
& harsh-fast & \cellcolor{green}0.285 & 21.529 & 1.322 & \cellcolor{green}0.098 \\
& harsh-slow & 0.229 & \cellcolor{green}0.021 & 1.193 & \cellcolor{green}0.085 \\
& low-fast & 0.242 & \cellcolor{green}0.056 & 1.159 & \cellcolor{green}0.058 \\
& low-slow & 0.186 & \cellcolor{green}0.024 & 1.231 & \cellcolor{green}0.106 \\
& neutral-fast & 0.210 & \cellcolor{green}0.049 & 1.285 & \cellcolor{green}0.053 \\
& neutral-slow & 0.228 & \cellcolor{green}0.012 & 1.307 & \cellcolor{green}0.110 \\
\Block{6-1}{orbit}
& harsh-fast & 0.243 & \cellcolor{green}0.021 & 1.107 & \cellcolor{green}0.055 \\
& harsh-slow & \cellcolor{green}0.280 & 79.796 & 1.098 & \cellcolor{green}0.143 \\
& low-fast & 0.253 & \cellcolor{green}0.030 & 1.063 & \cellcolor{green}0.045 \\
& low-slow & 0.271 & \cellcolor{green}0.028 & 1.048 & \cellcolor{green}0.061 \\
& neutral-fast & 0.249 & \cellcolor{green}0.019 & 0.996 & \cellcolor{green}0.039 \\
& neutral-slow & 0.224 & \cellcolor{green}0.031 & 1.008 & \cellcolor{green}0.224 \\
\end{NiceTabular}
\caption{Comparison of our pipeline $\bM_{\zeta}$ and SPNv2.}
\vspace{-2em}
\label{tab:results_spnv2_vs_ours}
\end{table}

We compared the pose estimation performance of our pipeline against SPNv2 proposed by Park~\etal~\cite{park2023spnv2} which amalgamates additional concepts like multi-task learning~\cite{wang2022bridging} and test time adaptation (TTA)~\cite{wang2021tent}. We trained both SPNv2 and our pipeline $\bM_{\zeta}$ on \texttt{synthetic} event data and performed TTA as well as inference on real event data. $\Phi$ and $\Psi$ are computed for all the real event scenes for SPNv2 and $\bM_{\zeta}$ and summarised in Tab.~\ref{tab:results_spnv2_vs_ours}. Although, the mean was skewed due to outlying event frames in $2$ scenes however in general these results demonstrated the superior performance of $\bM_{\zeta}$ for permutations of laboratory settings and SPNv2 which was designed for RGB data does not cope well with real event data. 
\vspace{-0.5em}
\section{Conclusions}
We captured a novel synthetic dataset consisting of RGB and event frames as well as accompanying ground-truth poses extensively covering the expected orientations of the satellite model for training. A new event  and RGB dataset was also captured in real-life lab settings with several new conditions to extend our previous work in \cite{jawaid2023towards}. These datasets were evaluated on the same base pipeline to conclusively demonstrate the lower Sim2Real domain gap of event sensors as compared to RGB sensors for the task of satellite pose estimation. Moreover, we proposed a test-time certifiable self-supervision scheme to further close the domain gap for event frames. A comparison against established TTA method indicated superior performance of our pipeline.

\bibliographystyle{ieeetran}
\bibliography{root}

\begin{thebibliography}{10}
\providecommand{\url}[1]{#1}
\csname url@rmstyle\endcsname
\providecommand{\newblock}{\relax}
\providecommand{\bibinfo}[2]{#2}
\providecommand\BIBentrySTDinterwordspacing{\spaceskip=0pt\relax}
\providecommand\BIBentryALTinterwordstretchfactor{4}
\providecommand\BIBentryALTinterwordspacing{\spaceskip=\fontdimen2\font plus
\BIBentryALTinterwordstretchfactor\fontdimen3\font minus \fontdimen4\font\relax}
\providecommand\BIBforeignlanguage[2]{{%
\expandafter\ifx\csname l@#1\endcsname\relax
\typeout{** WARNING: IEEEtran.bst: No hyphenation pattern has been}%
\typeout{** loaded for the language `#1'. Using the pattern for}%
\typeout{** the default language instead.}%
\else
\language=\csname l@#1\endcsname
\fi
#2}}

\bibitem{Cavaciuti2022in-space}
A.~J. Cavaciuti, J.~H. Heying, and J.~Davis, ``In-space servicing, assembly, and manufacturing for the new space economy,'' Center for Space Policy and Strategy, Aerospace Corporation, Tech. Rep., 2022.

\bibitem{PASQUALETTOCASSINIS2019100548}
L.~{Pasqualetto Cassinis}, R.~Fonod, and E.~Gill, ``Review of the robustness and applicability of monocular pose estimation systems for relative navigation with an uncooperative spacecraft,'' \emph{Progress in Aerospace Sciences}, vol. 110, p. 100548, 2019.

\bibitem{sharma2019spn}
S.~Sharma and S.~D'Amico, ``Pose estimation for non-cooperative rendezvous using neural networks,'' \emph{arXiv:1906.09868}, 2019.

\bibitem{bospec2019}
B.~Chen, J.~Cao, A.~Parra, and T.-J. Chin, ``Satellite pose estimation with deep landmark regression and nonlinear pose refinement,'' in \emph{ICCV Workshops}, 2019.

\bibitem{wang2022bridging}
Z.~Wang, M.~Chen, Y.~Guo, Z.~Li, and Q.~Yu, ``Bridging the domain gap in satellite pose estimation: a self-training approach based on geometrical constraints,'' \emph{CoRR}, vol. abs/2212.12103, 2022.

\bibitem{park2023spnv2}
T.~H. Park and S.~D’Amico, ``Robust multi-task learning and online refinement for spacecraft pose estimation across domain gap,'' \emph{Advances in Space Research}, 2023.

\bibitem{pauly2023survey}
L.~Pauly, W.~Rharbaoui, C.~Shneider, A.~Rathinam, V.~Gaudillière, and D.~Aouada, ``A survey on deep learning-based monocular spacecraft pose estimation: Current state, limitations and prospects,'' \emph{Acta Astronautica}, vol. 212, pp. 339--360, 2023.

\bibitem{proencca2020urso}
P.~F. Proen{\c{c}}a and Y.~Gao, ``Deep learning for spacecraft pose estimation from photorealistic rendering,'' in \emph{ICRA}.\hskip 1em plus 0.5em minus 0.4em\relax IEEE, 2020.

\bibitem{sparkchallenge21}
``Spark challenge: Spacecraft recognition leveraging knowledge of space environment,'' https://cvi2.uni.lu/spark-2021/.

\bibitem{park2021speedplus}
T.~H. Park, M.~M{\"a}rtens, G.~Lecuyer, D.~Izzo, and S.~D'Amico, ``{SPEED}+: Next-generation dataset for spacecraft pose estimation across domain gap,'' \emph{arXiv preprint arXiv:2110.03101}, 2021.

\bibitem{beierle19variable}
C.~Beierle and S.~D’Amico, ``Variable-magnification optical stimulator for training and validation of spaceborne vision-based navigation,'' \emph{Journal of Spacecraft and Rockets}, vol.~56, pp. 1060--1072, 2019.

\bibitem{WANG2018135}
M.~Wang and W.~Deng, ``Deep visual domain adaptation: A survey,'' \emph{Neurocomputing}, vol. 312, pp. 135--153, 2018.

\bibitem{posch2010qvga}
C.~Posch, D.~Matolin, and R.~Wohlgenannt, ``A qvga 143 db dynamic range frame-free pwm image sensor with lossless pixel-level video compression and time-domain cds,'' \emph{IEEE Journal of Solid-State Circuits}, vol.~46, no.~1, pp. 259--275, 2010.

\bibitem{vidal2018ultimate}
A.~R. Vidal, H.~Rebecq, T.~Horstschaefer, and D.~Scaramuzza, ``Ultimate slam? combining events, images, and imu for robust visual slam in hdr and high-speed scenarios,'' \emph{IEEE Robotics and Automation Letters}, vol.~3, no.~2, pp. 994--1001, 2018.

\bibitem{rebecq2019high}
H.~Rebecq, R.~Ranftl, V.~Koltun, and D.~Scaramuzza, ``High speed and high dynamic range video with an event camera,'' \emph{IEEE TPAMI}, vol.~43, no.~6, pp. 1964--1980, 2019.

\bibitem{izzo2022neuromorphic}
D.~Izzo, A.~Hadjiivanov, D.~Dold, G.~Meoni, and E.~Blazquez, ``Neuromorphic computing and sensing in space,'' \emph{arXiv preprint arXiv:2212.05236}, 2022.

\bibitem{mahlknecht2022exploring}
F.~Mahlknecht, D.~Gehrig, J.~Nash, F.~M. Rockenbauer, B.~Morrell, J.~Delaune, and D.~Scaramuzza, ``Exploring event camera-based odometry for planetary robots,'' \emph{arXiv preprint arXiv:2204.05880}, 2022.

\bibitem{sofiamcleod2022eccv}
S.~McLeod, G.~Meoni, D.~Izzo, A.~Mergy, D.~Liu, Y.~Latif, I.~Reid, and T.-J. Chin, ``Globally optimal event-based divergence estimation for ventral landing,'' in \emph{ECCV Workshops}, 2022.

\bibitem{jawaid2023towards}
M.~Jawaid, E.~Elms, Y.~Latif, and T.-J. Chin, ``Towards bridging the space domain gap for satellite pose estimation using event sensing,'' in \emph{ICRA}, 2023.

\bibitem{talak2023certifiable}
R.~Talak, L.~R. Peng, and L.~Carlone, ``Certifiable object pose estimation: Foundations, learning models, and self-training,'' \emph{{IEEE} Trans. Robotics}, vol.~39, no.~4, pp. 2805--2824, 2023.

\bibitem{shi2023correct}
J.~Shi, R.~Talak, D.~Maggio, and L.~Carlone, ``A correct-and-certify approach to self-supervise object pose estimators via ensemble self-training,'' in \emph{Robotics: Science and Systems}, 2023.

\bibitem{wang2021tent}
D.~Wang, E.~Shelhamer, S.~Liu, B.~A. Olshausen, and T.~Darrell, ``Tent: Fully test-time adaptation by entropy minimization,'' in \emph{ICLR}, 2021.

\bibitem{fehse2014rendezvous}
W.~Fehse, ``Rendezvous with and capture / removal of non-cooperative bodies in orbit: The technical challenges,'' \emph{Journal of Space Safety Engineering}, vol.~1, no.~1, pp. 17--27, 2014.

\bibitem{goddard2010onorbit}
NASA, ``On-orbit satellite servicing study,'' Goddard Space Flight Center, NASA, Tech. Rep., October 2010.

\bibitem{speed2020}
M.~Kisantal, S.~Sharma, T.~H. Park, D.~Izzo, M.~M{\"a}rtens, and S.~D’Amico, ``Satellite pose estimation challenge: Dataset, competition design, and results,'' \emph{IEEE TAES}, vol.~56, no.~5, 2020.

\bibitem{spec19}
``Kelvins satellite pose estimation competition 2019,'' https://kelvins.esa.int/satellite-pose-estimation-challenge/challenge/.

\bibitem{spec21}
\BIBentryALTinterwordspacing
``Kelvins satellite pose estimation competition 2021.'' [Online]. Available: \url{https://kelvins.esa.int/pose-estimation-2021/challenge/}
\BIBentrySTDinterwordspacing

\bibitem{sharma2018pose}
S.~Sharma, C.~Beierle, and S.~D'Amico, ``Pose estimation for non-cooperative spacecraft rendezvous using convolutional neural networks,'' in \emph{2018 IEEE Aerospace Conference}.\hskip 1em plus 0.5em minus 0.4em\relax IEEE, 2018, pp. 1--12.

\bibitem{huo2020fast}
Y.~Huo, Z.~Li, and F.~Zhang, ``Fast and accurate spacecraft pose estimation from single shot space imagery using box reliability and keypoints existence judgments,'' \emph{IEEE Access}, vol.~8, 2020.

\bibitem{wang2020self6d}
G.~Wang, F.~Manhardt, J.~Shao, X.~Ji, N.~Navab, and F.~Tombari, ``Self6d: Self-supervised monocular 6d object pose estimation,'' in \emph{ECCV}, vol. 12346, 2020, pp. 108--125.

\bibitem{wang2021occlusion}
G.~Wang, F.~Manhardt, X.~Liu, X.~Ji, and F.~Tombari, ``Occlusion-aware self-supervised monocular 6d object pose estimation,'' \emph{IEEE Transactions on Pattern Analysis and Machine Intelligence}, 2021.

\bibitem{chen2022sim}
K.~Chen, R.~Cao, S.~James, Y.~Li, Y.~Liu, P.~Abbeel, and Q.~Dou, ``Sim-to-real 6d object pose estimation via iterative self-training for robotic bin picking,'' in \emph{ECCV 2022}, vol. 13699, 2022, pp. 533--550.

\bibitem{wang2019normalized}
H.~Wang, S.~Sridhar, J.~Huang, J.~Valentin, S.~Song, and L.~J. Guibas, ``Normalized object coordinate space for category-level 6d object pose and size estimation,'' in \emph{CVPR 2019}, 2019.

\bibitem{zakharov2020autolabeling}
S.~Zakharov, W.~Kehl, A.~Bhargava, and A.~Gaidon, ``Autolabeling 3d objects with differentiable rendering of {SDF} shape priors,'' in \emph{CVPR}, 2020.

\bibitem{zhang2023self}
K.~Zhang, Y.~Fu, S.~Borse, H.~Cai, F.~Porikli, and X.~Wang, ``Self-supervised geometric correspondence for category-level 6d object pose estimation in the wild,'' in \emph{ICLR}, 2023.

\bibitem{deng2020self}
X.~Deng, Y.~Xiang, A.~Mousavian, C.~Eppner, T.~Bretl, and D.~Fox, ``Self-supervised 6d object pose estimation for robot manipulation,'' in \emph{ICRA 2020}.\hskip 1em plus 0.5em minus 0.4em\relax {IEEE}, 2020, pp. 3665--3671.

\bibitem{chin2019star}
T.-J. Chin, S.~Bagchi, A.~Eriksson, and A.~Van~Schaik, ``Star tracking using an event camera,'' in \emph{CVPR Workshops}, 2019.

\bibitem{ng2022asynchronous}
Y.~Ng, Y.~Latif, T.~Chin, and R.~E. Mahony, ``Asynchronous kalman filter for event-based star tracking,'' in \emph{ECCV Workshops}, 2022.

\bibitem{sikorski2021event}
O.~Sikorski, D.~Izzo, and G.~Meoni, ``Event-based spacecraft landing using time-to-contact,'' in \emph{CVPR 2021}, 2021, pp. 1941--1950.

\bibitem{azzalini2023on}
L.~J. Azzalini, E.~Blazquez, A.~Hadjiivanov, G.~Meoni, and D.~Izzo, ``On the generation of a synthetic event-based vision dataset for navigation and landing,'' \emph{CoRR}, vol. abs/2308.00394, 2023.

\bibitem{elms2022seenic}
\BIBentryALTinterwordspacing
E.~Elms, M.~Jawaid, Y.~Latif, and T.-J. Chin, ``{SEENIC: dataset for Spacecraft posE Estimation with NeuromorphIC vision},'' Oct. 2022. [Online]. Available: \url{https://doi.org/10.5281/zenodo.7214231}
\BIBentrySTDinterwordspacing

\bibitem{rathinam2023spades}
A.~Rathinam, H.~Qadadri, and D.~Aouada, ``Spades: A realistic spacecraft pose estimation dataset using event sensing,'' \emph{arXiv preprint arXiv:2311.05310}, 2023.

\bibitem{wu2019detectron2}
Y.~Wu, A.~Kirillov, F.~Massa, W.-Y. Lo, and R.~Girshick, ``Detectron2,'' \url{https://github.com/facebookresearch/detectron2}, 2019.

\bibitem{sun2019hrnet}
K.~Sun, B.~Xiao, D.~Liu, and J.~Wang, ``Deep high-resolution representation learning for human pose estimation,'' in \emph{CVPR}, 2019.

\bibitem{chen2020bpnp}
B.~Chen, A.~Parra, J.~Cao, N.~Li, and T.-J. Chin, ``End-to-end learnable geometric vision by backpropagating pnp optimization,'' in \emph{CVPR}, 2020.

\bibitem{blender}
\url{https://www.blender.org/}.

\bibitem{v2e2012katz}
M.~L. Katz, K.~Nikolic, and T.~Delbruck, ``Live demonstration: Behavioural emulation of event-based vision sensors,'' in \emph{ISCAS}, 2012.

\bibitem{gracca2023shining}
R.~Gra{\c{c}}a, B.~McReynolds, and T.~Delbruck, ``Shining light on the {DVS} pixel: A tutorial and discussion about biasing and optimization,'' in \emph{CVPR}, 2023.

\bibitem{MetavisionSDKDocs}
``Metavision sdk docs,'' https://docs.prophesee.ai/stable/index.html.

\bibitem{opencv_library}
G.~Bradski, ``{The OpenCV Library},'' 2000.

\bibitem{hu2021v2e}
Y.~Hu, S.-C. Liu, and T.~Delbruck, ``v2e: From video frames to realistic dvs events,'' in \emph{CVPR}, 2021, pp. 1312--1321.

\end{thebibliography}

\end{document}